\pdfoutput=1

\documentclass[11pt]{article}

\usepackage[]{emnlp2021}

\usepackage{times}
\usepackage{latexsym}

\usepackage[T1]{fontenc}

\usepackage[utf8]{inputenc}

\usepackage{microtype}

\usepackage{latexsym}
\usepackage{booktabs} 
\usepackage{multirow}
\usepackage{amsmath}
\usepackage{amssymb}
\usepackage{graphicx}
\usepackage{tabularx}
\usepackage{float}

\usepackage{natbib}

\usepackage{cleveref}
\crefformat{section}{\S#2#1#3}
\crefformat{subsection}{\S#2#1#3}
\crefformat{subsubsection}{\S#2#1#3}
\crefrangeformat{section}{\S\S#3#1#4 to~#5#2#6}
\crefmultiformat{section}{\S\S#2#1#3}{ and~#2#1#3}{, #2#1#3}{ and~#2#1#3}
\usepackage{refstyle}
\Crefformat{figure}{#2Fig.~#1#3}
\Crefmultiformat{figure}{Figs.~#2#1#3}{ and~#2#1#3}{, #2#1#3}{ and~#2#1#3}
\Crefformat{table}{#2Tab.~#1#3}
\Crefmultiformat{table}{Tabs.~#2#1#3}{ and~#2#1#3}{, #2#1#3}{ and~#2#1#3}
\Crefformat{equation}{#2Eq.~(#1#3)}

\Crefformat{appendix}{#2App.~\S#1#3}

\usepackage{colortbl}

\usepackage{arydshln}
\usepackage{caption}
\usepackage{subcaption}
\usepackage{xcolor}
\usepackage{microtype}

\usepackage{pifont}
\newcommand{\cmark}{\ding{51}}%
\newcommand{\xmark}{\ding{55}}%

\usepackage{hyperref}
\usepackage{tablefootnote}

\usepackage{xcolor,colortbl}
\definecolor{plot_orange}{rgb}{1.0, 0.498, 0.055}
\definecolor{plot_blue}{rgb}{0.121, 0.467, 0.706}
\definecolor{plot_grey}{rgb}{0.501, 0.501, 0.501}
\definecolor{plot_greyblue}{rgb}{0.475, 0.561, 0.725} 
\definecolor{plot_yellow}{rgb}{0.953, 0.659, 0.231} 

\usepackage{xspace}
\newcommand{\en}{{\textsc{en}}\xspace}

\newcommand{\es}{{\textsc{es}}\xspace}
\newcommand{\zh}{{\textsc{zh}}\xspace}
\newcommand{\tr}{{\textsc{tr}}\xspace}
\newcommand{\fin}{{\textsc{fi}}\xspace}
\newcommand{\ru}{{\textsc{ru}}\xspace}
\newcommand{\fr}{{\textsc{fr}}\xspace}

\newcommand{\ita}{{\textsc{it}}\xspace}

\newcommand{\et}{{\textsc{et}}\xspace}
\newcommand{\pl}{{\textsc{pl}}\xspace}
\newcommand{\ar}{{\textsc{ar}}\xspace}
\newcommand{\he}{{\textsc{he}}\xspace}

\title{Fast, Effective, and Self-Supervised: Transforming Masked Language Models into Universal Lexical and Sentence Encoders}

\author{Fangyu Liu, Ivan Vuli\'{c}, Anna Korhonen, Nigel Collier \\
Language Technology Lab, TAL, University of Cambridge\\
\texttt{\{fl399, iv250, alk23, nhc30\}cam.ac.uk} }

\begin{document}
\maketitle
\begin{abstract}
Previous work has indicated that pretrained Masked Language Models (MLMs) are not effective as universal lexical and sentence encoders off-the-shelf, i.e., without further task-specific fine-tuning on NLI, sentence similarity, or paraphrasing tasks using annotated task data. In this work, we demonstrate that it is possible to turn MLMs into effective lexical and sentence encoders even without any additional data, relying simply on self-supervision. We propose an extremely simple, fast, and effective contrastive learning technique, termed Mirror-BERT, which converts MLMs (e.g., BERT and RoBERTa) into such encoders in 20--30 seconds with no access to additional external knowledge. Mirror-BERT relies on identical \textit{and} slightly modified string pairs as positive (i.e., synonymous) fine-tuning examples, and aims to maximise their similarity during \textit{``identity fine-tuning''}. We report huge gains over off-the-shelf MLMs with Mirror-BERT both in lexical-level and in sentence-level tasks, across different domains and different languages. Notably, in sentence similarity (STS) and question-answer entailment (QNLI) tasks, our self-supervised Mirror-BERT model even matches the performance of the Sentence-BERT models from prior work which rely on annotated task data. Finally, we delve deeper into the inner workings of MLMs, and suggest some evidence on why this simple Mirror-BERT fine-tuning approach can yield effective universal lexical and sentence encoders.
\end{abstract}

\section{Introduction}
Transfer learning with pretrained Masked Language Models (MLMs) such as BERT \citep{devlin2019bert} and RoBERTa \citep{liu2019roberta} has been widely successful in NLP, offering unmatched performance in a large number of tasks \cite{Wang:2019nips}. Despite the wealth of semantic knowledge stored in the MLMs \cite{Rogers:2020tacl}, they do not produce high-quality lexical and sentence embeddings when used off-the-shelf, without further task-specific fine-tuning \citep{feng2020language,li-etal-2020-sentence}. In fact, previous work has shown that their performance is sometimes even below static word embeddings and specialised sentence encoders \cite{Cer:2018arxiv} in lexical and sentence-level semantic similarity tasks \citep{reimers2019sentence,vulic-etal-2020-probing,Litschko:2021ecir}.

\begin{figure}[t!]
    \centering
    \includegraphics[width=0.75\linewidth]{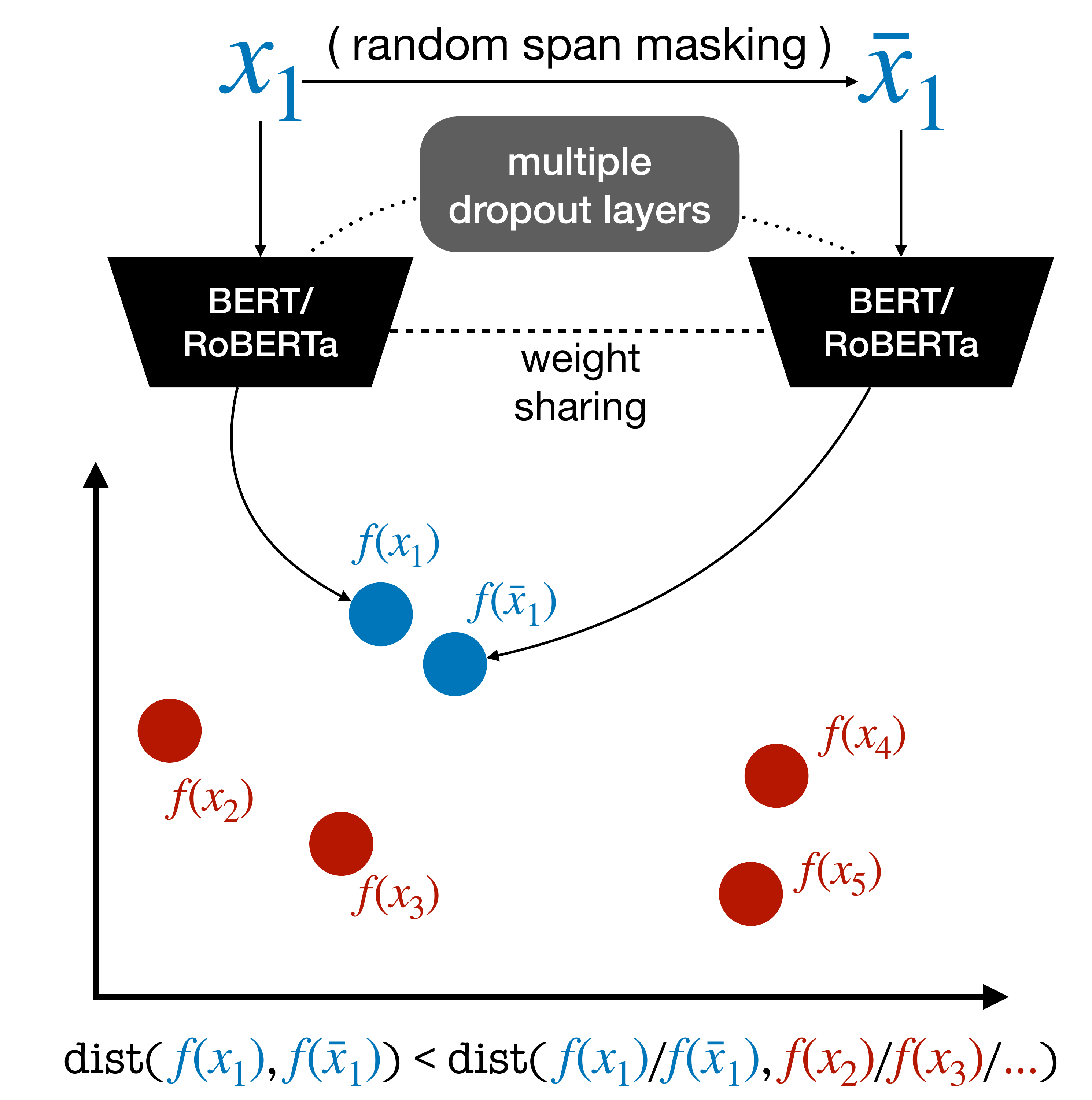}
    \caption{Illustration of the main concepts behind the proposed self-supervised Mirror-BERT method. The same text sequence can be observed from two additional ``views'': 1) by performing random span masking in the input space, and/or 2) by applying dropout (inside the BERT/RoBERTa MLM) in the feature space, yielding identity-based (i.e., ``mirrored'') positive examples for fine-tuning. A contrastive learning objective is then applied to encourage such ``mirrored'' positive pairs to obtain more similar representations in the embedding space relatively to negative pairs.}
    \label{fig:sketch_of_concept}
\end{figure}

In order to address this gap, recent work has trained dual-encoder networks on labelled external resources to convert MLMs into universal language encoders. Most notably, Sentence-BERT (SBERT, \citealt{reimers2019sentence}) further trains BERT and RoBERTa on Natural Language Inference (NLI, \citealt{bowman-etal-2015-large,Williams:2018naacl}) and sentence similarity data \cite{cer2017semeval} to obtain high-quality universal sentence embeddings. Recently, SapBERT \citep{liu2020self} self-aligns phrasal representations of the same meaning using synonyms extracted from the UMLS \cite{bodenreider2004unified}, a large biomedical knowledge base, obtaining lexical embeddings in the biomedical domain that reach state-of-the-art (SotA) performance in biomedical entity linking tasks. However, both SBERT and SapBERT require annotated (i.e., human-labelled) data as external knowledge: it is used to instruct the model to produce similar representations for text sequences (e.g., words, phrases, sentences) of similar/identical meanings.

In this paper, we fully dispose of any external supervision, demonstrating that the transformation of MLMs into universal language encoders can be achieved without task-labelled data. We propose a fine-tuning framework termed \textit{Mirror-BERT}, which simply relies on duplicating and slightly augmenting the existing text input (or their representations) to achieve the transformation, and show that it is possible to learn universal lexical and sentence encoders with such ``mirrored'' input data through self-supervision (see \Cref{fig:sketch_of_concept}). The proposed Mirror-BERT framework is also extremely efficient: the whole MLM transformation can be completed in less than one minute on two 2080Ti GPUs. 

Our findings further confirm a general hypothesis from prior work \cite{liu2020self,Zaken:2020bitfit,Glavas:2021eacl} that fine-tuning exposes the wealth of (semantic) knowledge stored in the MLMs.
In this case in particular, we demonstrate that the Mirror-BERT procedure can rewire the MLMs to serve as universal language encoders even without any external supervision. We further show that data augmentation in both input space and feature space are key to the success of Mirror-BERT, and they provide a synergistic effect. 

\vspace{1.5mm}
\noindent \textbf{Contributions.} \textbf{1)} We propose a completely self-supervised approach that can quickly transform pretrained MLMs into capable universal lexical and sentence encoders, greatly outperforming off-the-shelf MLMs in similarity tasks across different languages and domains. \textbf{2)} We investigate the rationales behind why Mirror-BERT works at all, aiming to understand the impact of data augmentation in the input space as well as in the feature space. We release our code and models at {\small \url{https://github.com/cambridgeltl/mirror-bert}}.

\section{Mirror-BERT: Methodology}
\label{sec:method}

Mirror-BERT consists of three main parts, described in what follows. First, we create positive pairs by duplicating the input text (\Cref{sec:self-dup}). We then further process the positive pairs by simple data augmentation operating either on the input text or on the feature map inside the model (\Cref{sec:data_aug}). Finally, we apply standard contrastive learning, `attracting' the texts belonging to the same class (i.e., positives) while pushing away the negatives (\Cref{sec:loss}).


\subsection{Training Data through Self-Duplication}\label{sec:self-dup}
The key to success of dual-network representation learning \cite[\textit{inter alia}]{Henderson:2019acl,reimers2019sentence,Humeau:2020iclr,liu2020self} is the construction of positive and negative pairs. While negative pairs can be easily obtained from randomly sampled texts, positive pairs usually need to be manually annotated. In practice, they are extracted from labelled task data (e.g., NLI) or knowledge bases that store relations such as synonymy or hypernymy (e.g., PPDB, \citealt{PPDB2}; BabelNet, \citealt{Ehrmann:14}; WordNet, \citealt{Fellbaum:1998wn}; UMLS). 

Mirror-BERT, however, does not rely on any external data to construct the positive examples. In a nutshell, given a set of non-duplicated strings $\mathcal{X}$, we assign individual labels ($y_i$) to each string and build a dataset $\mathcal{D}=\{(x_i, y_i) |x_i\in\mathcal{X}, y_i \in\{1,\ldots,|\mathcal{X}|\}\}$. We then create self-duplicated training data $\mathcal{D'}$ simply by repeating every element in $\mathcal{D}$.  
In other words, let $\mathcal{X} = \{x_1, x_2, \ldots\}$. We then have $\mathcal{D} = \{(x_1, y_1), (x_2, y_2), \ldots\}$ and $\mathcal{D'} = \{(x_1, y_1), (\overline{x}_1, \overline{y}_1), (x_2, y_2), (\overline{x}_2, \overline{y}_2), \ldots\}$ where $x_1 = \overline{x}_1, y_1 = \overline{y}_1, x_2 = \overline{x}_2, y_2 = \overline{y}_2, \ldots$. In \S\ref{sec:data_aug}, we introduce data augmentation techniques (in both input space and feature space) applied on $\mathcal{D'}$. Each positive pair $(x_i,\overline{x}_i)$ yields two different points/vectors in the encoder's representation space (see again \Cref{fig:sketch_of_concept}), and the distance between these points should be minimised.

\subsection{Data Augmentation}\label{sec:data_aug}
We hypothesise that applying certain `corruption' techniques to (i) parts of input text sequences or (ii) to their representations, or even (iii) doing both in combination, does not change their (captured) meaning. We present two `corruption' techniques as illustrated in \Cref{fig:sketch_of_concept}. First, we can directly mask parts of the input text. Second, we can erase (i.e., dropout) parts of their feature maps. Both techniques are rather simple and intuitive: (i) even when masking parts of an input sentence, humans can usually reconstruct its semantics; (ii) dropping a small subset of neurons or representation dimensions, the representations of a neural network will not drift too much.



\begin{figure}
    \centering
    \includegraphics[width=0.95\linewidth]{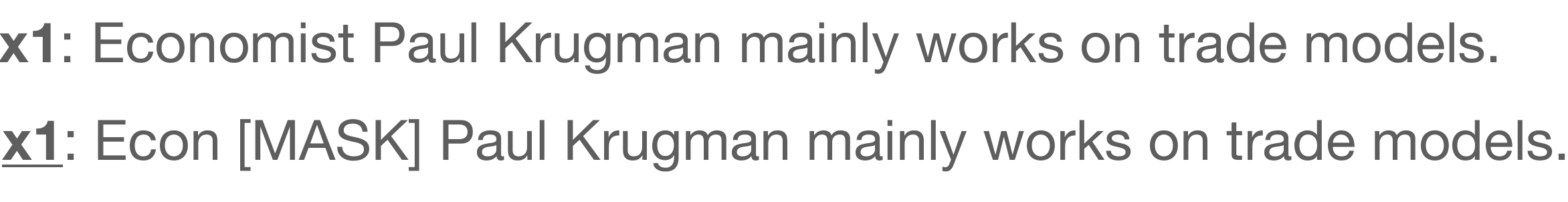}
    \vspace{-1mm}
    \caption{An example of input data augmentation via random span masking.}
    \label{fig:random_mask_aug}
\end{figure}

\vspace{1.5mm}
\noindent \textbf{Input Augmentation: Random Span Masking.} The idea is inspired by random cropping in visual representation learning \cite{Hendrycks:2020augmix}. In particular, starting from the mirrored pairs $(x_i, y_i)$ and $(\overline{x}_i, \overline{y}_i)$, we randomly replace a consecutive string of length $k$ with \texttt{[MASK]} in either $x_i$ or $\overline{x}_i$. The example (\Cref{fig:random_mask_aug}) illustrates the random span masking procedure with $k=5$.

\vspace{1.5mm}
\noindent \textbf{Feature Augmentation: Dropout.} 
The random span masking technique, operating directly on text input, can be applied only with sentence/phrase-level input; word-level tasks involve only short strings, usually represented as a single token under the sentence-piece tokeniser. However, data augmentation in the feature space based on dropout, as introduced below, can be applied to any input text. 

Dropout \citep{srivastava2014dropout} randomly drops neurons from a neural net during training with a probability $p$. In practice, it results in the erasure of each element with a probability of $p$. It has mostly been interpreted as implicitly bagging a large number of neural networks which share parameters at test time \citep{bouthillier2015dropout}. Here, we take advantage of the dropout layers in BERT/RoBERTa to create augmented views of the input text. Given a pair of identical strings $x_i$ and $\overline{x}_i$, their representations in the embedding space slightly differ due to the existence of multiple dropout layers in the BERT/RoBERTa architecture (\Cref{fig:dropout_aug}). The two data points in the embedding space can be seen as two augmented views of the same text sequence, which can be leveraged for fine-tuning.\footnote{The dropout augmentations are naturally a part of the BERT/RoBERTa network. That is, no further actions need to be taken to implement them. Note that random span masking is applied on only one side of the positive pair while dropout is applied on all data points.}


\begin{figure}
    \centering
    \includegraphics[width=0.87\linewidth]{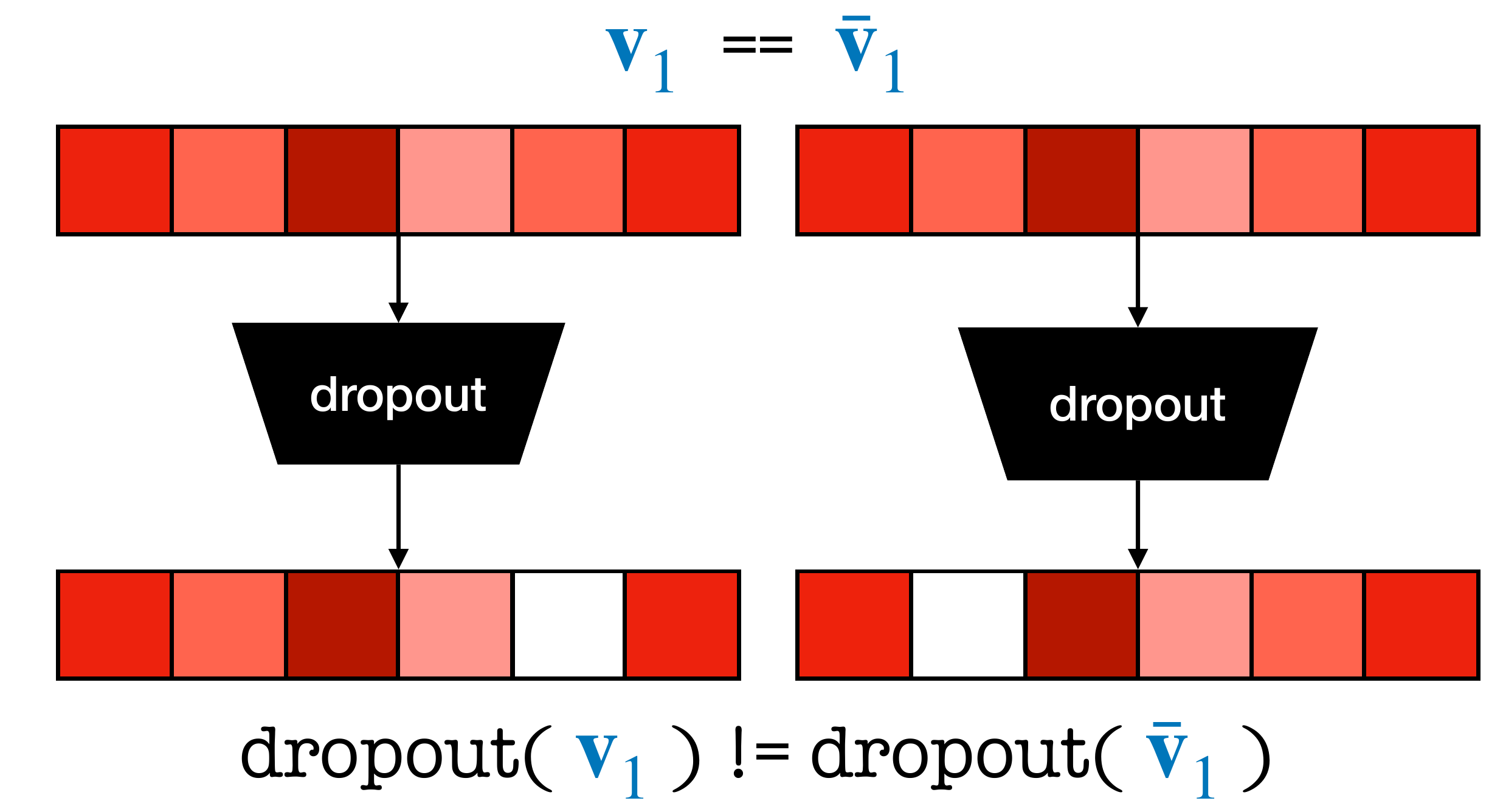}
    \caption{As the same vector goes through the same dropout layer separately, the outcomes are independent. Consequently, two fully identical strings fed to the single BERT/RoBERTa model yield different representations in the MLM embedding space.}
    \label{fig:dropout_aug}
\end{figure}

It is possible to combine data augmentation via random span masking and featuure augmentation via dropout; this variant is also evaluated later. 


\subsection{Contrastive Learning}\label{sec:loss}
Let $f(\cdot)$ denote the encoder model. The encoder is then fine-tuned on the data constructed in \S\ref{sec:data_aug}. Given a batch of data $\mathcal{D'}_b$, we leverage the standard InfoNCE loss \citep{oord2018representation} to cluster/attract the positive pairs together and push away the negative pairs in the embedding space:

\vspace{-2mm}
{\footnotesize
\begin{equation}
    \mathcal{L}_b = -\sum_{i=1}^{|\mathcal{D}_b|}\log\frac{\exp(\cos(f(x_i), f(\overline{x}_i))/\tau)}{\displaystyle \sum_{x_j\in \mathcal{N}_i}\exp(\cos(f(x_i), f(x_j))/\tau)}.
    \label{eq:infonce}
\end{equation}}%
$\tau$ denotes a temperature parameter; $\mathcal{N}_i$ denotes all negatives of $x_i$, which includes all $x_j,\overline{x}_j$ where $i\neq j$ in the current data batch (i.e., $|\mathcal{N}_i| = |\mathcal{D'}_b| - 2$). Intuitively, the numerator is the similarity of the self-duplicated pair (the positive example) and the denominator is the sum of the similarities between $x_i$ and all other strings besides $\overline{x}_i$ (the negatives).\footnote{We also experimented with another state-of-the-art contrastive learning scheme proposed by \citet{liu2020self}. There, hard triplet mining combined with multi-similarity loss (MS loss) is used as the learning objective. InfoNCE and triplet mining + MS loss work mostly on par, with slight gains of one variant in some tasks, and vice versa. For simplicity and brevity, we report the results only with InfoNCE.}

\section{Experimental Setup}
\label{sec:exp}


\noindent \textbf{Evaluation Tasks: Lexical.}
We evaluate on domain-general and domain-specific tasks: word similarity and biomedical entity linking (BEL). For the former, we rely on the Multi-SimLex evaluation set \citep{vulic2020multi}: it contains human-elicited word similarity scores for multiple languages. For the latter, we use NCBI-disease (NCBI, \citealt{dougan2014ncbi}), BC5CDR-disease, BC5CDR-chemical (BC5-d, BC5-c, \citealt{li2016biocreative}), AskAPatient \citep{limsopatham2016normalising} and COMETA (stratified-general split, \citealt{cometa}) as our evaluation datasets. The first three datasets are in the scientific domain (i.e., the data have been extracted from scientific papers), while the latter two are in the social media domain (i.e., extracted from online forums discussing health-related topics). We report Spearman's rank correlation coefficients ($\rho$) for word similarity; accuracy $@1/@5$ is the standard evaluation measure in the BEL task.

\vspace{1.5mm}
\noindent \textbf{Evaluation Tasks: Sentence-Level.}
Evaluation on the intrinsic sentence textual similarity (STS) task is conducted on the standard SemEval 2012-2016 datasets \citep{agirre2012semeval,agirre2013sem,agirre2014semeval,agirre2015semeval,agirre2016semeval}, STS Benchmark (STS-b, \citealt{cer2017semeval}), SICK-Relatedness (SICK-R, \citealt{marelli2014sick}) for English; STS SemEval-17 data is used for Spanish and Arabic \cite{cer2017semeval}, and we also evaluate on Russian STS.\footnote{\url{github.com/deepmipt/deepPavlovEval}} We report Spearman's $\rho$ rank correlation. Evaluation in the question-answer entailment task is conducted on QNLI \citep{rajpurkar2016squad,wang2018glue}. It contains 110k English QA pairs with binary entailment labels.\footnote{We follow the setup of \citet{li-etal-2020-sentence} and adapt QNLI to an unsupervised task by computing the AUC scores (on the development set, $\approx$5.4k pairs) using 0/1 labels and cosine similarity scores of QA embeddings.}




\vspace{1.5mm}
\noindent \textbf{Evaluation Tasks: Cross-Lingual.} We also assess the benefits of Mirror-BERT on cross-lingual representation learning, evaluating on cross-lingual word similarity (CLWS, Multi-SimLex is used) and bilingual lexicon induction (BLI). We rely on the standard mapping-based BLI setup \cite{artetxe-etal-2018-robust}, and training and test sets from \citet{glavas-etal-2019-properly}, reporting accuracy $@1$ scores (with CSLS as the word retrieval method, \citealt{lample2018word}).

\vspace{1.5mm}
\noindent \textbf{Mirror-BERT: Training Resources.} For fine-tuning (general-domain) lexical representations, we use the top 10k most frequent words in each language. For biomedical name representations, we randomly sample 10k names from the UMLS. In sentence-level tasks, for STS, we sample 10k sentences (without labels) from the training set of the STS Benchmark; for Spanish, Arabic and Russian, we sample 10k sentences from the WikiMatrix dataset \citep{schwenk2019wikimatrix}. For QNLI, we sample 10k sentences from its training set. 



\vspace{1.5mm}
\noindent \textbf{Training Setup and Details.} 
The hyperparameters of word-level models are tuned on SimLex-999 \citep{hill2015simlex}; biomedical models are tuned on COMETA (zero-shot-general split). Sentence-level models are tuned on the dev set of STS-b. $\tau$ in \Cref{eq:infonce} is $0.04$ (biomedical and sentence-level models); $0.2$ (word-level). Dropout rate $p$ is $0.1$. Sentence-level models use a random span masking rate of $k=5$, while $k=2$ for biomedical phrase-level models; we do not employ span masking for word-level models (an analysis is in the Appendix). All lexical models are trained for 2 epochs, max token length is 25. Sentence-level models are trained for 1 epoch with a max sequence length of 50. 

All models use AdamW \citep{loshchilov2018decoupled} as the optimiser, with a learning rate of \texttt{2e-5}, batch size of 200 (400 after duplication). In all tasks, for all `Mirror-tuned' models, unless noted otherwise, we create final representations using \texttt{[CLS]}, instead of another common option: \textit{mean-pooling (mp)} over all token representations in the last layer \cite{reimers2019sentence}.\footnote{For `non-Mirrored' original MLMs, the results with \textit{mp} are reported instead; they produce much better results than using \texttt{[CLS]}; see the Appendix.} \footnote{All reported results are averages of three runs. In general, the training is very stable, with negligible fluctuations.}



\section{Results and Discussion}
\label{s:results}
\subsection{Lexical-Level Tasks}
\label{ss:lexical}

\begin{table}[!t] 
\centering
\setlength{\tabcolsep}{3.8pt}
\renewcommand{\arraystretch}{0.85}
\scriptsize
\begin{tabular}{lccccccccccc}
\cmidrule[1.0pt]{1-10}
lang.$\rightarrow$ & \en & \fr & \et & \ar & \zh & \ru & \es & \pl & avg. \\
\cmidrule[1.0pt]{1-10}
fastText & \underline{.528} & \underline{.560} & \textbf{.447} & \underline{.409} & .428 & \textbf{.435} & \textbf{.488} & \underline{.396} & \underline{.461} \\
\cmidrule[1.0pt]{1-10}
BERT & .267 & .020 &.106 & .220 & .398 & .202 & .177 & .217 & .201 \\
\rowcolor{blue!10}
+ Mirror  & \textbf{.556} & \textbf{.621} & \underline{.308} & \textbf{.538} & \textbf{.639} & \underline{.365} & .296 & \textbf{.444} & \textbf{.471} \\
\cmidrule[1.0pt]{1-10}
mBERT & .105 & .130 &.094 & .101 & .261 & .109 & .095 & .087 & .123 \\
\rowcolor{blue!10}
+ Mirror & .358 & .341 & .134 & .097 & \underline{.501} & .210 & \underline{.332} & .141 & .264 \\
\cmidrule[1.0pt]{1-10}
\end{tabular}
\caption{Word similarity evaluation on Multi-SimLex. ``BERT'' denotes monolingual BERT models in each language (see the Appendix). ``mBERT'' denotes multilingual BERT. \textbf{Bold} and \underline{underline} denote highest and second-highest scores per column, respectively.}
\label{tab:ws}
\end{table}

\begin{table}[t!]
\setlength{\tabcolsep}{1.5pt}
\renewcommand{\arraystretch}{0.85}
\centering
\scriptsize
\centering
\begin{tabular}{lccccccccccccccccccc}
\cmidrule[1.0pt]{1-18}
& & \multicolumn{8}{c}{scientific language}  & \multicolumn{5}{c}{social media language}\\
\cmidrule[1.0pt]{2-9} \cmidrule[1.0pt]{11-15}
 \multirow{2}{*}{\shortstack[l]{dataset$\rightarrow$\\model$\downarrow$}} & \multicolumn{2}{c}{NCBI} &  &\multicolumn{2}{c}{BC5-d} &  &\multicolumn{2}{c}{BC5-c} & & \multicolumn{2}{c}{AskAPatient} & & \multicolumn{2}{c}{COMETA}   \\
\cmidrule[1.5pt]{2-3}\cmidrule[1.5pt]{5-6}\cmidrule[1.5pt]{8-9}\cmidrule[1.5pt]{11-12}\cmidrule[1.5pt]{14-15}
 & $@1$ & $@5$ & & $@1$ & $@5$ & & $@1$ & $@5$ & & $@1$ & $@5$  & & $@1$ & $@5$ \\
\cmidrule[1.0pt]{1-20}
SapBERT & \textbf{.920} & \textbf{.956} & & \textbf{.935} & \textbf{.960} &  & \textbf{.965} & \textbf{.982} &  & \textbf{.705} & \textbf{.889} && \textbf{.659} & \textbf{.779} \\
\cmidrule[1.0pt]{1-18}
BERT & .676 & .770 & &.815 & .891 & & .798 & .912 & & .382 & .433 & & .404 & .477 \\
\rowcolor{blue!10}
+ Mirror & .872 & .921 & & .921 & .949 & & .957 & .971 && .555 & .695 && .547 & .647 \\
\cmidrule[1.0pt]{1-18}
PubMedBERT & .778 & .869 & & .890 & .938 & & .930 & .946 & & .425 & .496 & & .468 & .532  \\
\rowcolor{blue!10}
+ Mirror & \underline{.909} & \underline{.948} && \underline{.930} & \underline{.962} && \underline{.958} & \underline{.979} && \underline{.590} & .\underline{750} && \underline{.603} & \underline{.713} \\
\cmidrule[1.0pt]{1-18}
\end{tabular}
\caption{Biomedical entity linking (BEL) evaluation.}
\label{tab:bel}
\end{table}

\begin{table*}[!t] 
\centering
\small
\begin{tabular}{l ccccccccccc}
\cmidrule[1.0pt]{1-10}
model$\downarrow$, dataset$\rightarrow$  & STS12 & STS13 & STS14 & STS15 & STS16 & STS-b & SICK-R & avg.\\
\cmidrule[1.0pt]{1-10}
SBERT* & \textbf{.719} & .774 &.\textbf{742} & .799 & .747 & .774 & \textbf{.721} & \underline{.754} \\
\cmidrule[1.0pt]{1-10}
BERT-CLS & .215 & .321 & .213 & .379 & .442 & .203 & .427 & .314 \\
BERT-mp & .314 & .536 & .433 & .582 & .596 & .464 & .528 & .493 \\
\rowcolor{blue!10}
+ Mirror & .670	& .801 & .713 & .812 & .743	& .764	& .699	& .743 \\
\rowcolor{blue!10}
+ Mirror (drophead) & \underline{.691}	& .811	& .730	& \underline{.819}	& .757	& .780	& .691 & \underline{.754} \\
\cmidrule[1.0pt]{1-10}
RoBERTa-CLS & .090 & .327 & .210 &.338 & .388 & .317 & .355 & .289 \\
RoBERTa-mp & .134 & .126 & .124 & .203 & .224 & .129 & .320 & .180 \\
\rowcolor{blue!10}
+ Mirror & .646	& \underline{.818}	& .734	& .802	& \underline{.782}	& \underline{.787} & \underline{.703} & .753 \\
\rowcolor{blue!10}
+ Mirror (drophead) & .666	& \textbf{.827}	& \underline{.740} & \textbf{.824}	& \textbf{.797}	& \textbf{.796}	& .697	& \textbf{.764} \\
\bottomrule
\end{tabular}
\caption{English STS. *We were able to reproduce the scores reported in the original Sentence-BERT (SBERT, \citealt{reimers2019sentence}) paper. However, we found mean-pooling over all tokens (including padded ones) yield better performance (.754 vs .749). We thus report the stronger baseline.}
\label{tab:sts_sota_full}
\end{table*}

\begin{table}[!t]
\centering
\scriptsize
\def\arraystretch{0.8}
\begin{tabularx}{\linewidth}{lXXXXXXXXXXX}
\cmidrule[1.0pt]{1-10}
model$\downarrow$, lang.$\rightarrow$  & \es & \ar & \ru & avg. \\
\cmidrule[1.0pt]{1-10}
BERT & .599 & .455 & .552 & .533 \\
\rowcolor{blue!10}
+ Mirror & \underline{.709} & \textbf{.669} & \underline{.673} & \textbf{.684} \\
\cmidrule[1.0pt]{1-10}
 mBERT & .610 & .447 & .616 & .558 \\
\rowcolor{blue!10}
 + Mirror & \textbf{.755} & \underline{.594} & \textbf{.692}  & \underline{.680} \\
\cmidrule[1.0pt]{1-10}
\end{tabularx}
\caption{STS evaluation in other languages.}
\label{tab:sts_multillingual}
\end{table}


\noindent \textbf{Word Similarity (\Cref{tab:ws}).}
SotA static word embeddings such as fastText \citep{mikolov2018advances} typically outperform off-the-shelf MLMs on word similarity datasets  \cite{vulic2020multi}. However, our results demonstrate that the Mirror-BERT procedure indeed converts the MLMs into much stronger word encoders. The Multi-SimLex results on 8 languages from \Cref{tab:ws} suggest that the fine-tuned \textit{+Mirror} variant substantially improves the performance of base MLMs (both monolingual and multilingual ones), even beating fastText in 5 out of the 8 evaluation languages.\footnote{Language codes: see the Appendix for a full listing.}

We also observe that it is essential to have a strong base MLM. While Mirror-BERT does offer substantial performance gains with all base MLMs, the improvement is more pronounced when the base model is strong (e.g., \textsc{en}, \textsc{zh}).

\vspace{1.2mm}
\noindent \textbf{Biomedical Entity Linking (\Cref{tab:bel}).}
The goal of BEL is to map a biomedical name mention to a controlled vocabulary (usually a node in a knowledge graph). Considered a downstream application in BioNLP, the BEL task also helps evaluate and compare the quality of biomedical name representations: it requires pairwise comparisons between the biomedical mention and all surface strings stored in the biomedical knowledge graph.

The results from \Cref{tab:bel} suggest that our \textit{+Mirror} transformation achieves very strong gains on top of the base MLMs, both BERT and PubMedBERT \citep{pubmedbert}.
We note that PubMedBERT is a current SotA MLM in the biomedical domain, and performs significantly better than BERT, both before and after \textit{+Mirror} fine-tuning. This highlights the necessity of starting from a domain-specific model when possible. On scientific datasets, the self-supervised \textit{PubMedBERT+Mirror} model is very close to SapBERT, which fine-tunes PubMedBERT with more than 10 million synonyms extracted from the external UMLS knowledge base. 

However, in the social media domain, \textit{PubMedBERT+Mirror} still cannot match the performance of  knowledge-guided SapBERT. This in fact reflects the nature and complexity of the task domain. For the three datasets in the scientific domain (NCBI, BC5-d, BC5-c), strings with similar surface forms tend to be associated with the same concept. On the other hand, in the social media domain, semantics of very different surface strings might be the same.\footnote{For instance, \emph{HCQ} and \emph{Plaquenil} refer to exactly the same concept on online health forums: \emph{Hydroxychloroquine}.} This also suggests that Mirror-BERT adapts PubMedBERT to a very good surface-form encoder for biomedical names, but dealing with more difficult synonymy relations (e.g. as found in the social media) does need external knowledge.\footnote{Motivated by these insights, in future work we will also investigate a combined approach that blends self-supervision and external knowledge \citep{Vulic:2021lexfit}, which could also be automatically mined \citep{simbert,thakur-etal-2021-augmented}.}

\begin{figure}[t!]
    \centering
    \includegraphics[width=1.0\linewidth]{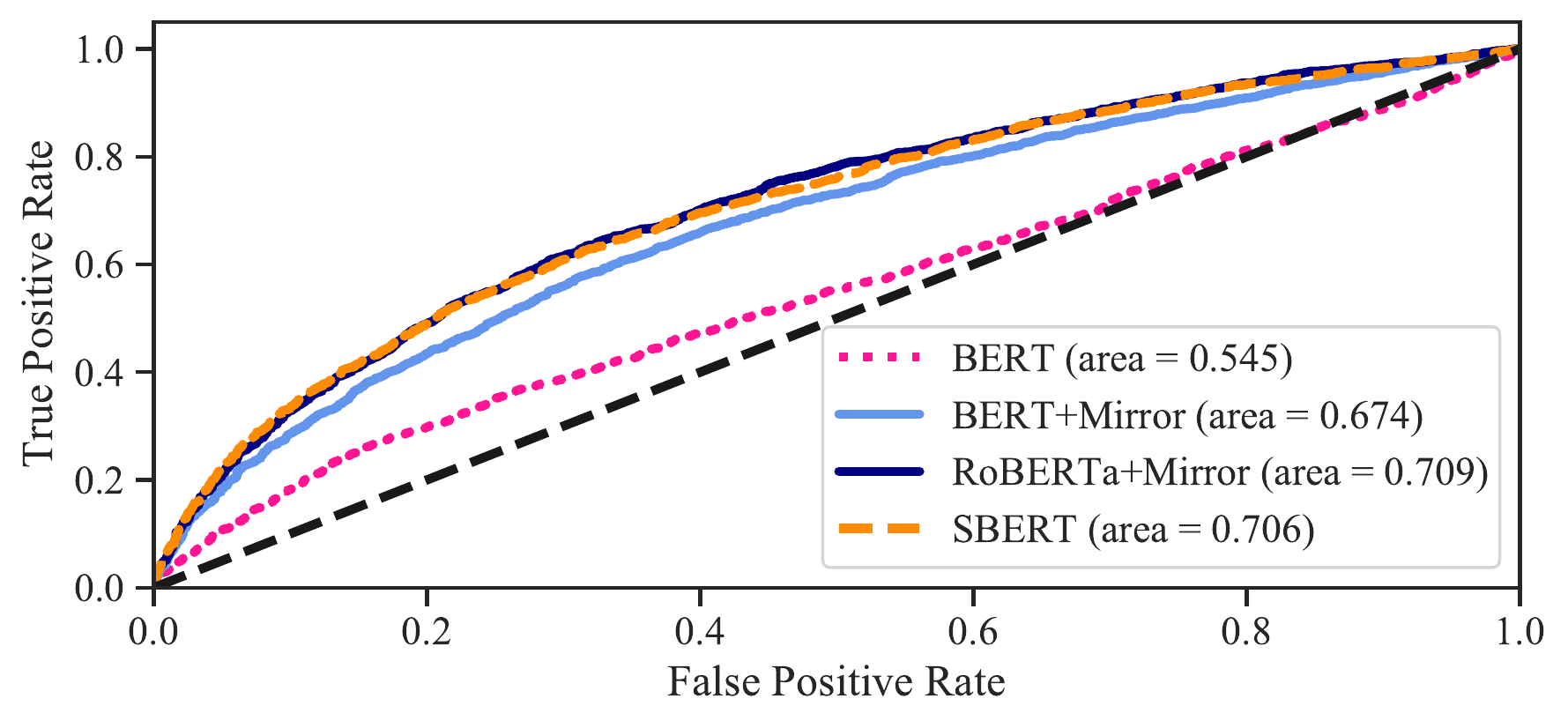}
    \caption{Unsupervised QNLI: ROC curves and AOC scores.}
    \label{fig:qnli}
\end{figure}
\begin{table}[!t] 
\centering
\setlength{\tabcolsep}{2.0pt}
\def\arraystretch{0.8}
\scriptsize
\begin{tabularx}{\linewidth}{lXXXXXXXXXXX}
\cmidrule[1.0pt]{1-10}
lang.$\rightarrow$ & \en-\fr & \en-\zh & \en-\he & \fr-\zh & \fr-\he & \zh-\he & avg.\\
\cmidrule[1.0pt]{1-10}
mBERT & .163 & .118 & .071 & .142 & .104 & .010 & .101\\
\rowcolor{blue!10}
+ Mirror  & \textbf{.454} & \textbf{.385} & \textbf{.133} & \textbf{.465} & \textbf{.163} & \textbf{.179} & \textbf{.297} \\
\cmidrule[1.0pt]{1-10}
\end{tabularx}
\caption{Cross-lingual word similarity results.}
\label{tab:ws_xling}
\end{table}
\begin{table}[!t] 
\centering
\setlength{\tabcolsep}{2.2pt}
\def\arraystretch{0.8}
\scriptsize
\begin{tabularx}{\linewidth}{lXXXXXXXXXXXX}
\cmidrule[1.0pt]{1-10}
lang.$\rightarrow$ & \en-\fr & \en-\ita & \en-\ru & \en-\tr & \ita-\fr & \ru-\fr & avg. \\
\cmidrule[1.0pt]{1-10}
BERT & .014 & .112 & .154 & .150 & .025 & .018 & .079 \\
\rowcolor{blue!10}
+ Mirror  & \textbf{.458} & \textbf{.378} & \textbf{.336} & \textbf{.289} & \textbf{.417} & \textbf{.345} & \textbf{.371} \\
\cmidrule[1.0pt]{1-10}
\end{tabularx}
\caption{BLI results.}
\label{tab:bli}
\end{table}


\subsection{Sentence-Level Tasks}\label{sec:sentence}


\noindent \textbf{English STS (\Cref{tab:sts_sota_full}).} Regardless of the base model (BERT/RoBERTa), applying  \textit{+Mirror} fine-tuning greatly boosts performance across all English STS datasets. Surprisingly, on average, \textit{RoBERTa+Mirror}, fine-tuned with only 10k sentences without any external supervision, is on-par with the SBERT model, which is trained on the merged SNLI \citep{bowman-etal-2015-large} and MultiNLI \citep{Williams:2018naacl} datasets, containing 570k and 430k sentence pairs, respectively.



\vspace{1.5mm}
\noindent \textbf{Spanish, Arabic and Russian STS (\Cref{tab:sts_multillingual}).} The results in the STS tasks on other languages, which all have different scripts, again indicate very large gains, using both monolingual language-specific BERTs and mBERT as base MLMs. This confirms that Mirror-BERT is a language-agnostic method.

\vspace{1.5mm}
\noindent \textbf{Question-Answer Entailment (\Cref{fig:qnli}).} The results indicate that our \textit{+Mirror} fine-tuning consistently improves the underlying MLMs. The \textit{RoBERTa+Mirror} variant even shows a slight edge over the supervised SBERT model (.709 vs. .706).



\subsection{Cross-Lingual Tasks}\label{sec:xling}


We observe huge gains across all language pairs in CLWS (\Cref{tab:ws_xling}) and BLI (\Cref{tab:bli}) after running the Mirror-BERT procedure. For language pairs that involve Hebrew, the improvement is usually smaller. We suspect that this is due to mBERT itself containing poor semantic knowledge for Hebrew. This finding aligns with our prior argument that a strong base MLM is still required to obtain prominent gains from running Mirror-BERT.  


\subsection{Further Discussion and Analyses}
\label{ss:further}

\noindent \textbf{Running Time.}
The Mirror-BERT procedure is extremely time-efficient. While fine-tuning on NLI (SBERT) or UMLS (SapBERT) data can take hours, Mirror-BERT with 10k positive pairs completes the conversion from MLMs to universal language encoders within a minute on two NVIDIA RTX 2080Ti GPUs. On average, 10-20 seconds is needed for 1 epoch of the Mirror-BERT procedure.

\begin{figure}[t!]
    \centering
    \includegraphics[width=0.94\linewidth]{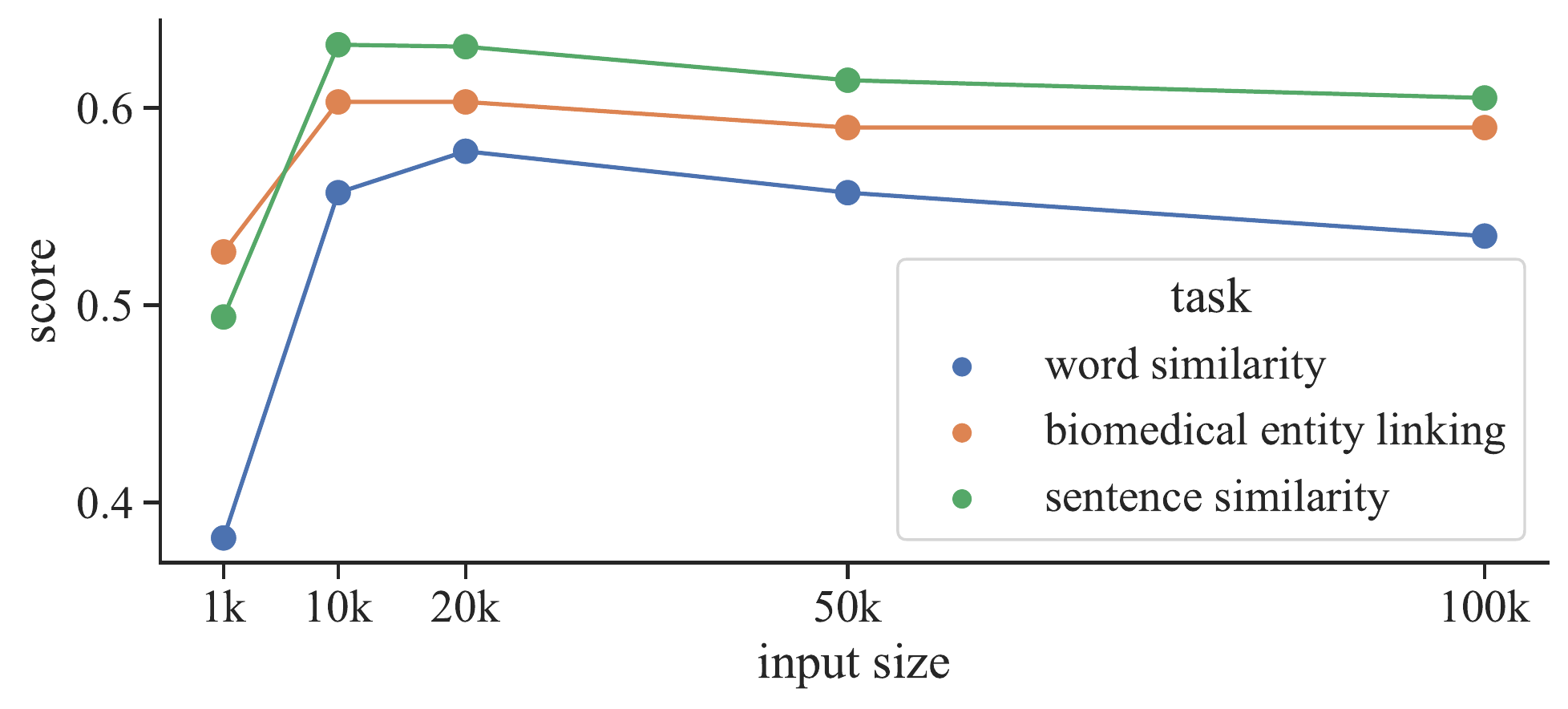}
    \caption{The impact of the number of fine-tuning ``mirrored'' examples ($x$-axis) on the task performance ($y$-axis). The scores across tasks are not directly comparable, and are based on different evaluation metrics (\Cref{sec:exp}).}
    \label{fig:input_size_ablation}
\end{figure}

\vspace{1.5mm}
\noindent \textbf{Input Data Size (\Cref{fig:input_size_ablation}).} In our main experiments in \Cref{ss:lexical}-\Cref{sec:xling}, we always use 10k examples for Mirror-BERT tuning. In order to assess the importance of the fine-tuning data size, we run a relevant analysis for a subset of base MLMs, and on a subset of English tasks. In particular, we evaluate the following: (i) BERT, Multi-SimLex (\en) (word-level); (ii) PubMedBERT, COMETA (biomedical phrase-level); (iii) RoBERTa, STS12 (sentence-level). The results indicate that the performance in all tasks reaches its peak in the region of 10k-20k examples and then gradually decreases, with a steeper drop on the the word-level task.\footnote{
We suspect that this is due to the inclusion of lower-frequency words into the fine-tuning data: embeddings of such words typically obtain less reliable embeddings \cite{Pilehvar:2018emnlp}.}
\footnote{For word-level experiments, we used the top 100k words in English according to Wikipedia statistics. For phrase-level experiments, we randomly sampled 100k names from UMLS. For sentence-level experiments we sampled 100k sentences from SNLI and MultiNLI datasets (as the STS training set has fewer than 100k sentences).}

\vspace{1.5mm}
\noindent \textbf{Random Span Masking + Dropout? (\Cref{tab:synersitic}).} 
We conduct our ablation studies on the English STS tasks. First, we experiment with turning off dropout, random span masking, or both. With both techniques turned off, we observe large performance drops of \textit{RoBERTa+Mirror} and \textit{BERT+Mirror} (see also the Appendix). Span masking appears to be the more important factor: its absence causes a larger decrease. However, the best performance is achieved when both dropout and random span masking are leveraged, suggesting a synergistic effect when the two augmentation techniques are used together. 



\begin{table}[!t]
\centering
\fontsize{10}{10}\selectfont
\def\arraystretch{0.9}
\begin{tabularx}{\linewidth}{XXXXX XXXXXXXXXXX}
\cmidrule[1.0pt]{1-10}
model configuration & avg. $\rho$\\
\cmidrule[1.0pt]{1-10}
RoBERTa + Mirror & .753 \\
\hdashline
- dropout + drophead & .764 ${\uparrow .011}$  \\
\hdashline
 - dropout & .732 ${\downarrow .021}$ \\
 - span mask & .717 ${\downarrow .036}$ \\
 - dropout \& span mask & .682 ${\downarrow .071}$ \\
\cmidrule[1.0pt]{1-10}
\end{tabularx}
\caption{Ablation study: (i) replacing dropout with drophead; (ii) the synergistic effect of dropout and random span masking in the English STS tasks.}
\label{tab:synersitic}
\end{table}

\vspace{1.5mm}
\noindent \textbf{Other Data Augmentation Types? Dropout vs. Drophead (\Cref{tab:synersitic}).}
Encouraged by the effectiveness of random span masking and dropout for Mirror-BERT, a natural question to pose is: can other augmentation types work as well? 
Recent work points out that pretrained MLM are heavily overparameterised and most Transformer heads can be pruned without hurting task performance \citep{voita2019analyzing,kovaleva2019revealing,michel2019sixteen}. \citet{zhou2020scheduled} propose a \textit{drophead} method: it randomly prunes attention heads at MLM training as a regularisation step. We thus evaluate a variant of Mirror-BERT where the dropout layers are replaced with such dropheads:\footnote{Drophead rates for BERT and RoBERTa are set to the default values of $0.2$ and $0.05$, respectively.} this results in even stronger STS performance, cf.~\Cref{tab:synersitic}. In short, this hints that the Mirror-BERT framework might benefit from other data and feature augmentation techniques in future work.\footnote{Besides the drophead-based feature space augmentation, in our side experiments, we also tested input space augmentation techniques such as whole-word masking, random token masking, and word reordering; they typically yield performance similar or worse to random span masking. We also point to very recent work that explores text augmentation in a different context \citep{wu2020clear,meng2021coco}. We leave a thorough search of optimal augmentation techniques for future work.}

\begin{figure}
    \centering
    \includegraphics[width=0.87\linewidth]{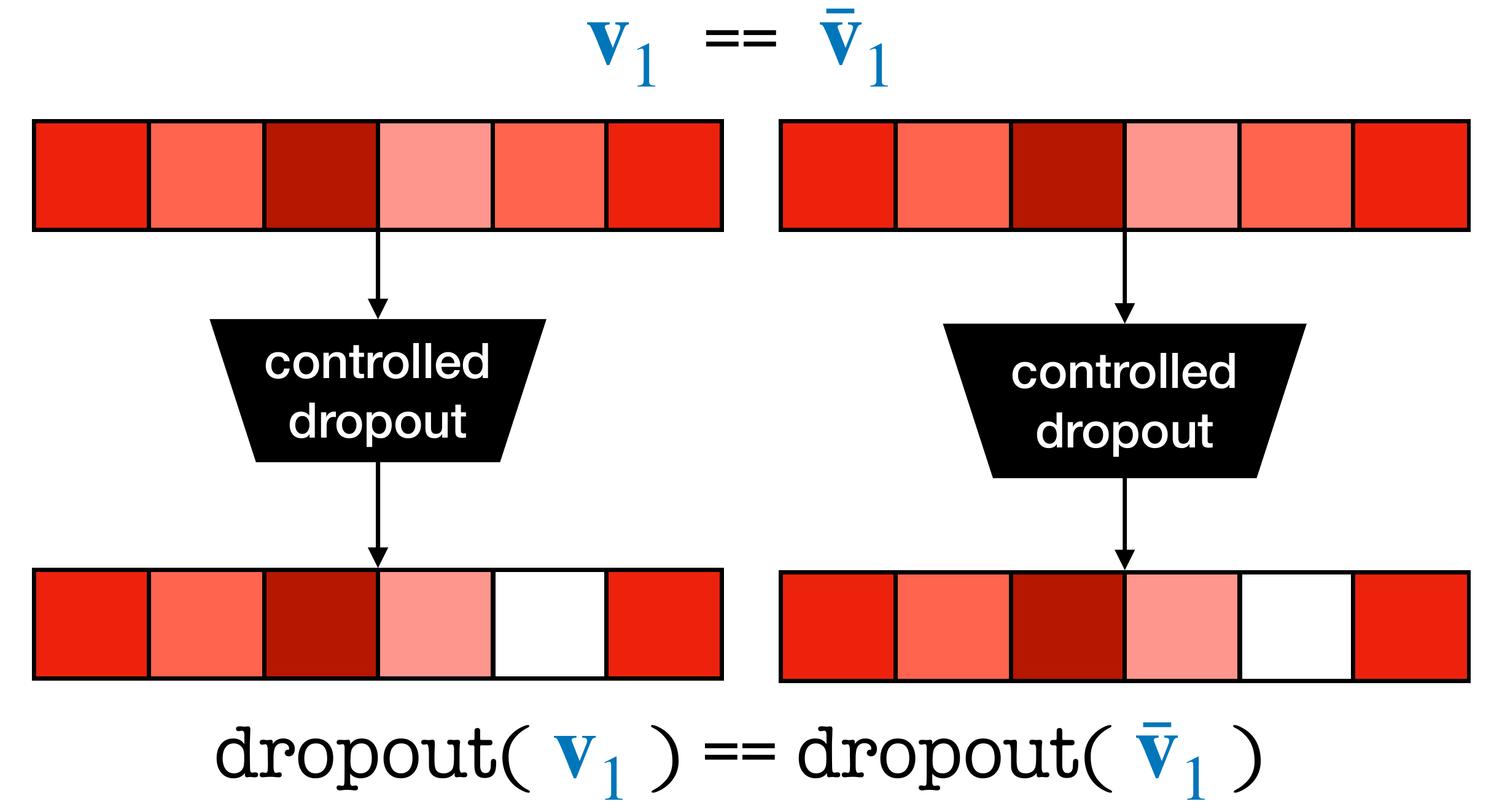}
    \vspace{-1mm}
    \caption{Under controlled dropout, if two strings are identical, they will have an identical set of dropout masks throughout the encoding process.}
    \label{fig:dropout_aug}
    \vspace{-1.0mm}
\end{figure}

\vspace{1.5mm}
\noindent \textbf{Regularisation or Augmentation? (\Cref{tab:dropout_probe}).} 
When using dropout, is it possible that we are simply observing the effect of adding/removing regularisation instead of the augmentation benefit? To answer this question, we design a simple probe that attempts to disentangle the effect of regularisation versus augmentation; we turn off random span masking but leave the dropout on (so that the regularisation effect remains). 
\begin{table}[]
\centering
\setlength{\tabcolsep}{2.0pt}
\def\arraystretch{0.8}
\fontsize{7.9}{10}\selectfont
\begin{tabularx}{\linewidth}{lXXXXXXXX}
\cmidrule[1.0pt]{1-5}
model configuration (MLM=RoBERTa) & $\rho$ on STS12 \\
\cmidrule[1.0pt]{1-5}
random span masking \xmark; dropout \xmark  & .562 \\
\hdashline
random span masking \xmark; dropout \cmark  & .648 ${\uparrow.086}$ \\
random span masking \xmark; controlled dropout \cmark  & .452 ${\downarrow.110}$ \\
\cmidrule[1.0pt]{1-5}
\end{tabularx}
\caption{Probing the impact of dropout.}
\label{tab:dropout_probe}
\end{table}
 However, instead of assigning independent dropouts to every individual string (rendering each string slightly different), we control the dropouts applied to a positive pair to be identical. As a result, it holds $f(x_i)=f(\overline{x}_i),$ when $x_i\equiv\overline{x}_i, \forall i\in\{1,\cdots,|\mathcal{D}|\}$. We denote this as ``controlled dropout''. In \Cref{tab:dropout_probe}, we observe that, during the \textit{+Mirror} fine-tuning, controlled dropout largely underperforms standard dropout and is even worse than not using dropout at all. As the only difference between controlled and standard dropout is the augmented features for positive pairs in the latter case, this suggests that the gain from \textit{+Mirror} indeed stems from the data augmentation effect rather than from regularisation. 

\begin{figure}[!t]
 \centering
 \subfloat[BERT-CLS]{\label{fig:cosine_distribution_a}\includegraphics[width=0.33\linewidth]{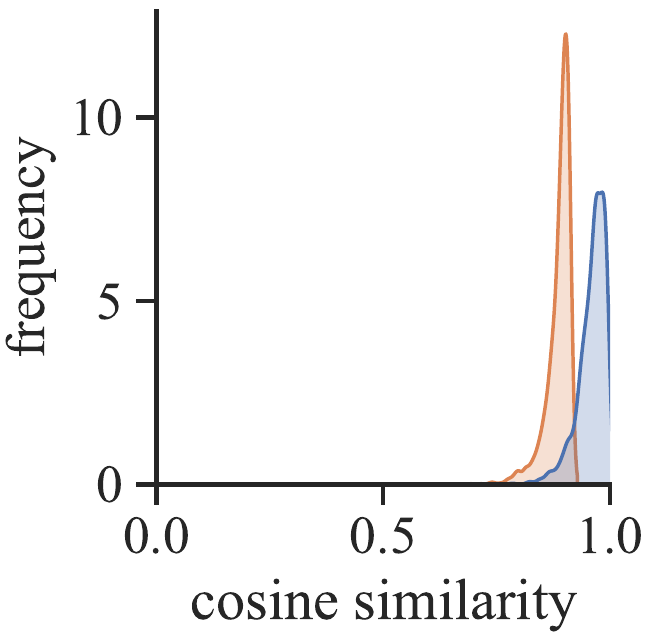}}\hfill
  \subfloat[BERT-mp]{\label{fig:cosine_distribution_b}\includegraphics[width=0.33\linewidth]{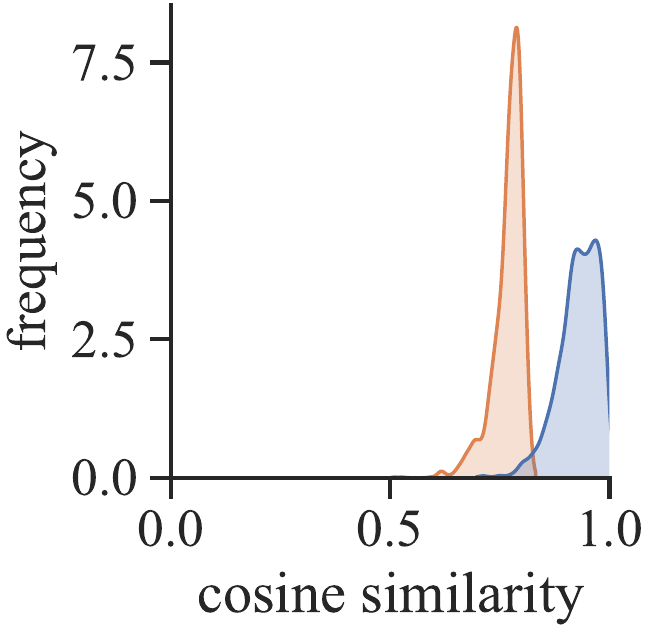}}%
\subfloat[BERT + Mirror]{\label{fig:cosine_distribution_c}\includegraphics[width=0.33\linewidth]{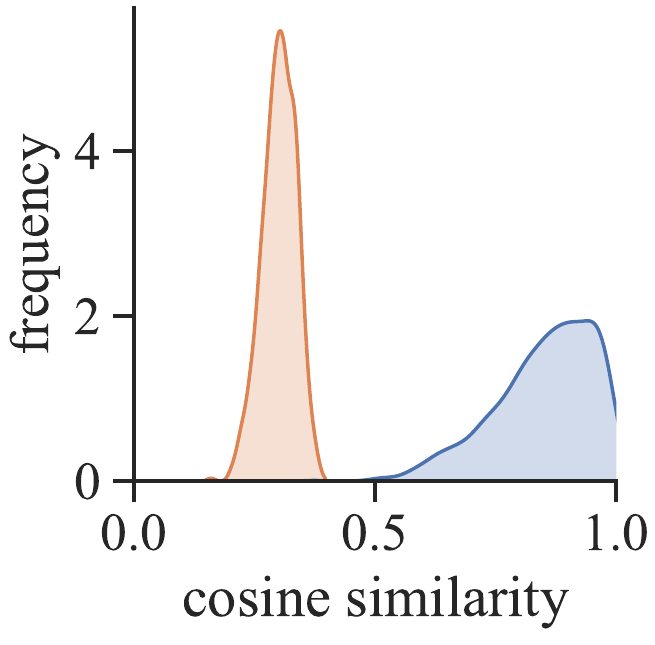}}%
\vspace{-1mm}
\caption{Cosine similarity distribution over 1k sentence pairs sampled from QQP. {\color{plot_blue}Blue} and {\color{plot_orange}orange} mean positive and negative similarities, respectively.}
\label{fig:cosine_distribution}
\vspace{-1.0mm}
\end{figure}

\vspace{1.2mm}
\noindent \textbf{Mirror-BERT Improves Isotropy? (\Cref{fig:cosine_distribution}).} We argue that the gains with Mirror-BERT largely stem from its reshaping of the embedding space geometry. \textit{Isotropy} (i.e., uniformity in all orientations) of the embedding space has been a favourable property for semantic similarity tasks \citep{arora2016latent,mu2017all}. However, \citet{ethayarajh2019contextual} shows that (off-the-shelf) MLMs' representations are anisotropic: they reside in a narrow cone in the vector space and the average cosine similarity of (random) data points is extremely high. Sentence embeddings induced from MLMs without fine-tuning thus suffer from spatial anistropy \citep{li-etal-2020-sentence,su2021whitening}. Is Mirror-BERT then improving isotropy of the embedding space?\footnote{Some preliminary evidence from \Cref{tab:synersitic} already leads in this direction: we observe large gains over the base MLMs even without any positive examples, that is, when both span masking and dropout are not used (i.e., it always holds $x_i = \overline{x}_i$ and $f(x_i) = f(\overline{x}_i)$). During training, this leads to a constant numerator in \Cref{eq:infonce}. In this case, learning collapses to the scenario where all gradients solely come from the negatives: the model is simply pushing all data points away from each other, resulting in a more isotropic space.} To investigate this claim, we inspect (1) the distributions of cosine similarities and (2) an isotropy score, as defined by \citet{mu2017all}. 


First, we randomly sample 1,000 sentence pairs from the Quora Question Pairs (QQP) dataset. In \Cref{fig:cosine_distribution}, we plot the distributions of pairwise cosine similarities of BERT representations before (\Cref{fig:cosine_distribution_a,fig:cosine_distribution_b}) and after the \textit{+Mirror} tuning (\Cref{fig:cosine_distribution_c}). The overall cosine similarities (regardless of positive/negative) are greatly reduced and the positives/negatives become easily separable. 


We also leverage a quantitative isotropy score (IS), proposed in prior work \citep{arora2016latent,mu2017all}, and defined as follows:

\vspace{-1.5mm}
{\footnotesize
\begin{equation}
  \text{IS} (\mathcal{V}) =  \frac{\min_{\mathbf{c}\in \mathcal{C}} \sum_{\mathbf{v}\in \mathcal{V}}\exp(\mathbf{c}^\top\mathbf{v})}{\max_{\mathbf{c}\in \mathcal{C}} \sum_{\mathbf{v}\in \mathcal{V}}\exp(\mathbf{c}^\top\mathbf{v})}
  \label{eq:isotropy_score}
\end{equation}
}%
where $\mathcal{V}$ is the set of vectors,\footnote{$\mathcal{V}$ comprises the corresponding text data used for Mirror-BERT fine-tuning (10k items for each task type).} $\mathcal{C}$ is the set of all possible unit vectors (i.e., any $\mathbf{c}$ so that $\|\mathbf{c}\|=1$) in the embedding space. In practice, $\mathcal{C}$ is approximated by the eigenvector set of $\mathbf{V}^\top\mathbf{V}$ ($\mathbf{V}$ is the stacked embeddings of $\mathcal{V}$). The larger the IS value, more isotropic an embedding space is (i.e., a perfectly isotropic space obtains the IS score of 1). 

IS scores in \Cref{tab:is} confirm that the \textit{+Mirror} fine-tuning indeed makes the embedding space more isotropic. Interestingly, with both data augmentation techniques switched off, a naive expectation is that IS will increase as the gradients now solely come from negative examples, pushing apart points in the space. However, we observe the increase of IS only for word-level representations. This hints at more complex dynamics between isotropy and gradients from positive and negative examples, where positives might also contribute to isotropy in some settings. We will examine these dynamics more in future work.\footnote{Introducing positive examples also naturally yields stronger task performance, as the original semantic space is better preserved. \citet{gao2021simcse} provide an insightful analysis on the balance of learning uniformity and alignment preservation, based on the method of \citet{wang2020understanding}.}



\begin{table}[!t]
\centering
\def\arraystretch{0.8}
\fontsize{7.9}{10}\selectfont
\begin{tabular}{lccccccccccc}
\cmidrule[1.0pt]{1-10}
level$\rightarrow$ & word & & phrase & & sentence \\
\midrule
BERT  & .169 && .205 && .222 \\
+ Mirror & .599 &&  \textbf{.252} && \textbf{.265} \\
+ Mirror (w/o aug.) &  \textbf{.825} && .170 && .255  \\
\cmidrule[1.0pt]{1-10}
\end{tabular}
\vspace{-1.5mm}
\caption{IS of word, phrase, and sentence-level models.}
\label{tab:is}
\vspace{-1mm}
\end{table}

\begin{table}[]
\centering
\setlength{\tabcolsep}{2.0pt}
\small
\begin{tabular}{lccccccccccc}
\cmidrule[1.0pt]{1-10}
model &  $\rho$ \\
\cmidrule[1.0pt]{1-10}
fastText & .528 \\
\cmidrule[1.0pt]{1-10}
BERT-CLS & .105 \\
BERT-mp & .267 \\
\hdashline
+ Mirror & .556 \\
\hdashline
+ Mirror (random string) & .393 \\
+ Mirror (random string, lr \texttt{5e-5}) & .481 \\
\cmidrule[1.0pt]{1-10}
\end{tabular}
 \vspace{-1.5mm}
\caption{Running Mirror-BERT with a set of `zero-semantics' random strings. Evaluation is conducted on Multi-SimLex (\en).}
\label{tab:random_string_full}
\end{table}

\vspace{1.5mm}
\noindent \textbf{Learning New Knowledge or Exposing Available Knowledge?} Running Mirror-BERT for more epochs, or with more data (see \Cref{fig:input_size_ablation}) does not result in performance gains. This hints that, instead of gaining new knowledge from the fine-tuning data, Mirror-BERT in fact `rewires' existing knowledge in MLMs \cite{Zaken:2020bitfit}. To further verify this, we run Mirror-BERT with random `zero-semantics' words, generated by uniformly sampling English letters and digits, and evaluate on (\en) Multi-SimLex. Surprisingly, even these data can transform off-the-shelf MLMs into effective word encoders: we observe a large improvement over the base MLM in this extreme scenario, from $\rho=$0.267 to 0.481 (\Cref{tab:random_string_full}). We did a similar experiment on the sentence-level and observed similar trends. However, we note that using the actual English texts for fine-tuning still performs better as they are more `in-domain' (with further evidence and discussions in the following paragraph). 

\begin{figure}[t!]
    \centering
    \includegraphics[width=1.0\linewidth]{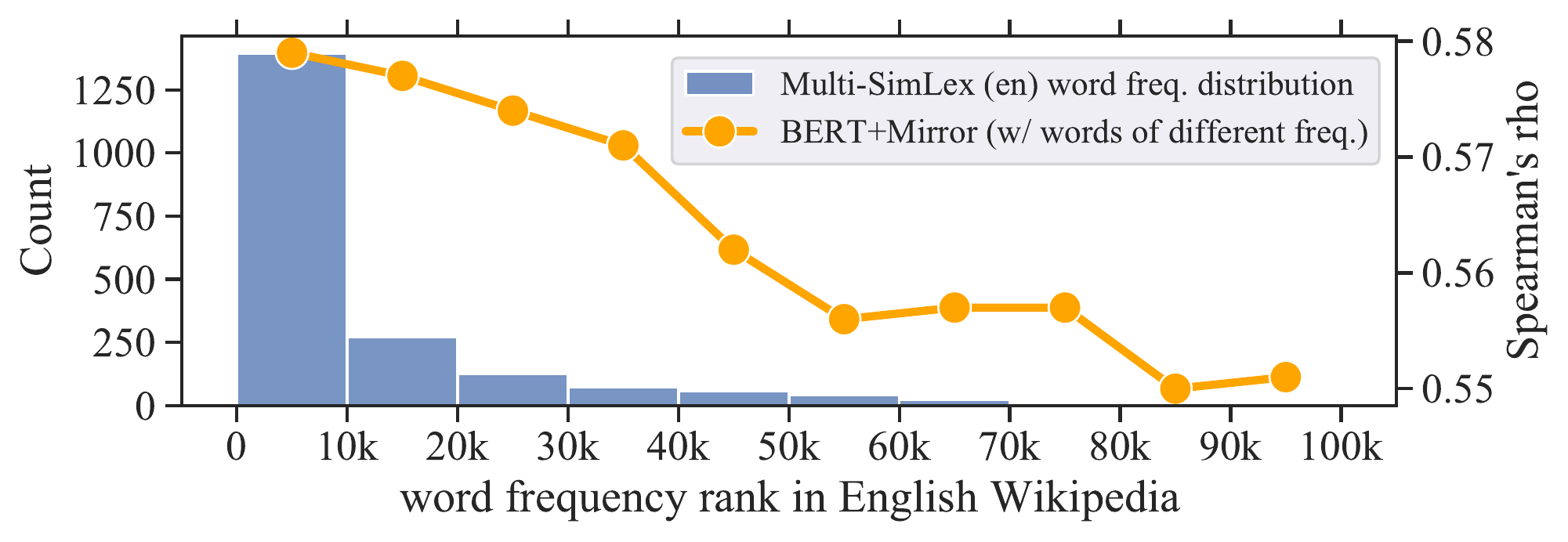}
    \vspace{-3.0mm}
    \caption{{\color{plot_greyblue}Blue}: words in Multi-SimLex (\en) follow a long-tail distribution. {\color{plot_yellow}Yellow}: \textit{BERT+Mirror} trained with frequent words tend to perform better.}
    \label{fig:word_freq}
\end{figure}







\vspace{1.5mm}
\noindent \textbf{Selecting Examples for Fine-Tuning.} Using raw text sequences from the end task should be the default option for Mirror-BERT fine-tuning since they are in-distribution by default, as semantic similarity models tend to underperform when faced with a domain shift \citep{zhang2020unsupervised}. In the \textit{general-domain} STS tasks, we find that using sentences extracted from the STS training set, Wikipedia articles, or NLI datasets all yield similar STS performance after running Mirror-BERT (though optimal hyperparameters differ). However, porting \textit{BERT+Mirror} trained on STS data to QNLI results in AUC drops from .674 to .665. This suggests that slight or large domain shifts do affect task performance, further corroborated by our findings from fine-tuning with fully random strings (see before). 



Further, \Cref{fig:word_freq} shows a clear tendency that more frequent strings are more likely to yield good task performance. There, we split the 100k most frequent words from English Wikipedia into 10 equally sized fine-tuning buckets of 10k examples each, and run \textit{+Mirror} fine-tuning on BERT with each bucket. In sum, using frequent in-domain examples seems to be the optimal choice.


\section{Related Work}
Self-supervised text representations have a large body of literature. Here, due to space constraints, we provide a highly condensed summary of the most related work. Even prior to the emergence of large pretrained LMs (PLMs), most representation models followed the distributional hypothesis \citep{harris1954distributional} and exploited the co-occurrence statistics of words/phrases/sentences in large corpora  \citep{mikolov2013efficient,mikolov2013distributed,pennington-etal-2014-glove,kiros2015skip,hill2016learning,logeswaran2018an}. Recently, DeCLUTR \citep{giorgi-etal-2021-declutr} follows the  distributional hypothesis and formulates sentence embedding training as a contrastive learning task where span pairs sampled from the same document are treated as positive pairs. Very recently, there has been a growing interest in using individual raw sentences for self-supervised contrastive learning on top of PLMs. 

\citet{wu2020clear} explore input augmentation techniques for sentence representation learning with contrastive objectives. However, they use it as an auxiliary loss during full-fledged MLM pretraining from scratch \citep{rethmeier2021primer}. In contrast, our post-hoc approach offers a lightweight and fast self-supervised transformation from any pretrained MLM to a universal language encoder at lexical or sentence level.

\citet{carlsson2021semantic} use two distinct models to produce two views of the same text, where we rely on a single model, that is, we propose to use dropout and random span masking within the same model to produce the two views, and demonstrate their synergistic effect. Our study also explores word-level and phrase-level representations and tasks, and to domain-specialised representations (e.g., for the BEL task).

SimCSE \citep{gao2021simcse}, a work concurrent to ours, adopts the same contrastive loss as Mirror-BERT, and also indicates the importance of data augmentation through dropout. However, they do not investigate random span masking as data augmentation in the input space, and limit their model to general-domain English sentence representations only, effectively rendering SimCSE a special case of the Mirror-BERT framework.
Other concurrent papers explore a similar idea, such as Self-Guided Contrastive Learning \citep{kim-etal-2021-self}, ConSERT \citep{yan-etal-2021-consert}, and BSL \citep{zhang-etal-2021-bootstrapped}, \textit{inter alia}. They all create two views of the same sentence for contrastive learning, with different strategies in feature extraction, data augmentation, model updating or choice of loss function. However, they offer less complete empirical findings compared to our work: we additionally evaluate on (1) lexical-level tasks, (2) tasks in a specialised biomedical domain and (3) cross-lingual tasks.

\section{Conclusion} \label{sec:conclusion}
We proposed Mirror-BERT, a simple, fast, self-supervised, and highly effective approach that transforms large pretrained masked language models (MLMs) into universal lexical and sentence encoders within a minute, and without any external supervision. Mirror-BERT, based on simple unsupervised data augmentation techniques, demonstrates surprisingly strong performance in (word-level and sentence-level) semantic similarity tasks, as well as on biomedical entity linking. The large gains over base MLMs are observed for different languages with different scripts, and across diverse domains. Moreover, we dissected and analysed the main causes behind Mirror-BERT's efficacy.


\section*{Acknowledgements}
We thank the reviewers and the AC for their considerate comments. We also thank the LTL members and Xun Wang for insightful feedback. FL is supported by Grace \& Thomas C.H. Chan Cambridge Scholarship. AK and IV are supported by the ERC Grant LEXICAL (no. 648909) and the ERC PoC Grant MultiConvAI (no. 957356). NC kindly acknowledges grant-in-aid funding from ESRC (grant number ES/T012277/1).
\bibliography{anthology,custom}
\bibliographystyle{acl_natbib}

\clearpage
\appendix

\section{Language Codes}\label{sec:appendix_lang_code}
\begin{table}[!h]
    \centering
    \begin{tabular}{ll}
    \toprule
         \en & English  \\
         \es & Spanish \\
         \fr & French \\
         \pl & Polish \\
         \et & Estonian \\
         \fin & Finnish \\
         \ru & Russian \\
         \tr & Turkish \\
         \ita & Italian \\
         \zh & Chinese \\
         \ar & Arabic \\
         \he & Hebrew \\
    \bottomrule
    \end{tabular}
    \caption{Language abbreviations used in the paper.}
  \label{tab:lang_code}
\end{table}

\section{Additional Training Details}
\paragraph{Most Frequent 10k/100k Words by Language.} The most frequent 10k words in each language were selected based on the following list: \\ \url{https://github.com/oprogramador/most-common-words-by-language}. 

\vspace{1.2mm}
\noindent The most frequent 100k English words in Wikipedia can be found here: \\
\url{https://gist.github.com/h3xx/1976236}.

\paragraph{\texttt{[CLS]} or Mean-Pooling?} For MLMs, the consensus in the community, also validated by our own experiments, is that mean-pooling performs better than using \texttt{[CLS]} as the final output representation. However, for Mirror-BERT models, we found \texttt{[CLS]} (before pooling) generally performs better than mean-pooling. The exception is BERT on sentence-level tasks, where we found mean-pooling performs better than \texttt{[CLS]}. In sum, sentence-level \textit{BERT+Mirror} models are fine-tuned and tested with mean-pooling while all other Mirror-BERT models are fine-tuned and tested with \texttt{[CLS]}. We also tried representations after the pooling layer, but found no improvement.

\paragraph{Training Stability.} All task results are reported as averages over three runs with different random seeds (if applicable). In general, fine-tuning is very stable and the fluctuations with different random seeds are very small. For instance, on the sentence-level task STS, the standard deviation is $<0.002$. On word-level, standard deviation is a bit higher, but is generally $<0.005$. Note that the randomly sampled training sets are fixed across all experiments, and changing the training corpus for each run might lead to larger fluctuations. 

\section{Details of \textit{Mirror-BERT} Trained on Random Strings}
We pointed out in the main text that \textit{BERT+Mirror} trained on random strings can outperform MLMs by large margins. 
With standard training configurations, BERT improves from .267 (BERT-mp) to .393 with \textit{+Mirror}. When learning rate is increased to \texttt{5e-5}, the MLM fine-tuned with random strings performs only around 0.07 lower than the standard \textit{BERT+Mirror} model fine-tuned with the 10k most frequent English words.

\section{Dropout and Random Span Masking Rates}

\noindent \textbf{Dropout Rate (\Cref{tab:dropout_rate}).} The performance trends conditioned on dropout rates are generally the same across word-level, phrase-level and sentence-level fine-tuning. Here, we use the STS task as a reference point. BERT prefers larger dropouts (0.2 \& 0.3) and is generally more robust. RoBERTa prefers a smaller dropout rate (0.05) and its performance decreases more steeply with the increase of the dropout rate. For simplicity, as mentioned in the main paper, we use the default value of 0.1 as the dropout rate for all models.

\begin{table}[]
\centering
\setlength{\tabcolsep}{2.0pt}
\small
\begin{tabularx}{\linewidth}{lXXXXXXXXXXX}
\cmidrule[1.0pt]{1-10}
dropout rate$\rightarrow$ &  0.05 & 0.1$^\ast$ & 0.2 & 0.3 & 0.4\\
\cmidrule[1.0pt]{1-10}
BERT + Mirror & .740 & .743 & \textbf{.748} & \textbf{.748} & .731 \\
RoBERTa + Mirror & \textbf{.755} & .753 & .737 & .694 & .677 \\
\bottomrule
\end{tabularx}
\caption{Average $\rho$ across STS tasks with different dropout rates. $^\ast$ default dropout rate for all models in other experiments.}
\label{tab:dropout_rate}
\end{table}

\vspace{1.3mm}
\noindent \textbf{Random Span Masking Rate (\Cref{tab:rm_rate}).} Interestingly, the opposite holds for random span masking: RoBERTa is more robust to larger masking rates $k$, and is much more robust than BERT to this hyper-parameter.

\begin{table}[]
\centering
\setlength{\tabcolsep}{2.0pt}
\small
\begin{tabularx}{\linewidth}{lXXXXXXXXXXX}
\cmidrule[1.0pt]{1-10}
random span mask rate$\rightarrow$  & 2 & 5$^\ast$ & 10 & 15 & 20 \\
\cmidrule[1.0pt]{1-10}
BERT + Mirror & .741 & \textbf{.743} & .720 & .690 & .616 \\
RoBERTa + Mirror & .750 & .753 & \textbf{.757} & .743 & .706  \\
\bottomrule
\end{tabularx}
\caption{Avg. $\rho$ across STS tasks with different random span masking rates. $^\ast$ default mask rates for all models in other experiments.}
\label{tab:rm_rate}
\end{table}

\section{Mean-Vector $l_2$-Norm (MVN)}
To supplement the quantitative evidence already suggested by the Isotropy Score (IS) in the main paper, we additionally compute the  mean-vector $l_2$-norm (MVN) of embeddings. In the word embedding literature, mean-centering has been a widely studied post-processing technique for inducing better semantic representations. \citet{mu2017all} point out that mean-centering is essentially increasing spatial isotropy by shifting the centre of the space to the region where actual data points reside in. Given a set of representation vectors $\mathcal{V}$, we define MVN as follows:
\begin{equation}
\text{MVN}(\mathcal{V}) =   \left\Vert \sum_{\mathbf{v}\in \mathcal{V}} \frac{\mathbf{v}}{  |\mathcal{V}|}\right\Vert_2.
\end{equation}
The lower MVN is, the more mean-centered an embedding is. As shown in \Cref{tab:is_full}, MVN aligns with the trends observed with IS. This further confirms our intuition that +\textit{Mirror} tuning makes the space more isotropic and shifts the centre of space close to the centre of data points. 

Very recently, \citet{cai2021isotropy} defined more metrics to measure spatial isotropy. \citet{rajaee-pilehvar-2021-cluster} also used \Cref{eq:isotropy_score} for analysing sentence embedding's isotropiness.

\begin{table}[]
\centering
\setlength{\tabcolsep}{2.0pt}
\small
\begin{tabular}{lccccccccccc}
\cmidrule[1.0pt]{1-10}
 \multirow{2}{*}{\shortstack[l]{level$\rightarrow$\\model$\downarrow$}} & \multicolumn{2}{c}{word} &  &\multicolumn{2}{c}{phrase} &  &\multicolumn{2}{c}{sentence} \\
\cmidrule[1.5pt]{2-3}\cmidrule[1.5pt]{5-6}\cmidrule[1.5pt]{8-9}
 & MVN & IS & & MVN & IS & & MVN & IS \\
\cmidrule[1.0pt]{1-10}
BERT-CLS & 13.79 & .043 && 12.8 & .028 && 12.73 & .062  \\
BERT-mp  & 7.89 & .169 && 6.82 & .205 && 6.93 & .222 \\
\hdashline
+ Mirror & 2.11 & .599 && \textbf{5.91} & \textbf{.252} && \textbf{5.57} & \textbf{.265} \\
+ Mirror (w/o aug.) & \textbf{0.71} & \textbf{.825} && 8.16 & .170 && 5.75 & .255  \\
\cmidrule[1.0pt]{1-10}
\end{tabular}
\caption{Full table for MVN and IS of word-, phrase-, and sentence-level models. Higher is better, that is, more isotropic with IS, while the opposite holds for MVN (lower scores mean more isotropic representation spaces).}
\label{tab:is_full}
\end{table}


\section{Evaluation Dataset Details}
All datasets used and links to download them can be found in the code repository provided. The Russian STS dataset is provided by \\
\url{https://github.com/deepmipt/deepPavlovEval}. The Quora Question Pair (QQP) dataset is downloaded at  \url{https://www.kaggle.com/c/quora-question-pairs}.

\section{Pretrained Encoders}\label{sec:appendix_huggingface_urls}
A complete listing of URLs for all used pretrained encoders is provided in \Cref{tab:model_url}. For monolingual MLMs of each language, we made the best effort to select the most popular one (based on download counts). For computational tractability of the large number of experiments conducted, all models are \textsc{Base} models (instead of \textsc{Large}).

\begin{table*}[h] 
\small
\setlength{\tabcolsep}{1pt}
\centering
\begin{tabular}{ll}
\toprule
model & URL \\
\midrule
fastText & \url{https://fasttext.cc/docs/en/crawl-vectors.html} \\
SBERT & \url{https://huggingface.co/sentence-transformers/bert-base-nli-mean-tokens} \\
SapBERT & \url{https://huggingface.co/cambridgeltl/SapBERT-from-PubMedBERT-fulltext} \\
BERT (English) & \url{https://huggingface.co/bert-base-uncased} \\
RoBERTa (English) & \url{https://huggingface.co/roberta-base} \\
mBERT & \url{https://huggingface.co/bert-base-multilingual-uncased} \\
Turkish BERT & \url{dbmdz/bert-base-turkish-uncased} \\
Italian BERT & \url{dbmdz/bert-base-italian-uncased} \\
French BERT & \url{https://huggingface.co/camembert-base} \\
Spanish BERT  & \url{https://huggingface.co/dccuchile/bert-base-spanish-wwm-uncased} \\
Russian BERT  &  \url{https://huggingface.co/DeepPavlov/rubert-base-cased} \\
Chinese BERT  &  \url{https://huggingface.co/bert-base-chinese} \\
Arabic BERT  &  \url{https://huggingface.co/aubmindlab/bert-base-arabertv02} \\
Polish BERT  &  \url{https://huggingface.co/dkleczek/bert-base-polish-uncased-v1} \\
Estonian BERT  &  \url{https://huggingface.co/tartuNLP/EstBERT} \\

\bottomrule
\end{tabular}
\caption{A listing of HuggingFace \& fastText URLs of all pretrained models used in this work.}
\label{tab:model_url}
\end{table*}

\section{Full Tables}

Here, we provide the complete sets of results. In these tables we include both MLMs w/ features extracted using both mean-pooling (``mp'') and \texttt{[CLS]} (``CLS'').

For full multilingual word similarity results, view \Cref{tab:ws_full}. 
For full Spanish, Arabic and Russian STS results, view \Cref{tab:sts_sota_full}. For full cross-lingual word similarity results, view \Cref{tab:ws_xling_full}. For full BLI results, view \Cref{tab:bli_full}. For full ablation study results, view \Cref{tab:synersitic_full}. For full MVN and IS scores, view \Cref{tab:is_full}. 

\begin{table*}[] 
\centering
\small
\begin{tabular}{lccccccccccc}
\cmidrule[1.0pt]{1-10}
language$\rightarrow$ & \en & \fr & \et & \ar & \zh & \ru & \es & \pl & avg. \\
\cmidrule[1.0pt]{1-10}
fastText & \underline{.528} & \underline{.560} & \textbf{.447} & \underline{.409} & .428 & \textbf{.435} & \textbf{.488} & \underline{.396} & \underline{.461} \\
\cmidrule[1.0pt]{1-10}
BERT-CLS & .105 & .050 &.160 & .210 & .277 & .177 & .152 & .257 & .174 \\
BERT-mp & .267 & .020 &.106 & .220 & .398 & .202 & .177 & .217 & .201 \\
\rowcolor{blue!10}
+ Mirror  & \textbf{.556} & \textbf{.621} & .308 & \textbf{.538} & \textbf{.639} & \underline{.365} & .296 & \textbf{.444} & \textbf{.471} \\
\cmidrule[1.0pt]{1-10}
mBERT-CLS & .062 & .046 & .074 & .047 & .204 & .063 & .039 & .051 & .073 \\
mBERT-mp & .105 & .130 &.094 & .101 & .261 & .109 & .095 & .087 & .123 \\
\rowcolor{blue!10}
+ Mirror & .358 & .341 & .134 & .097 & \underline{.501} & .210 & \underline{.332} & .141 & .264 \\
\cmidrule[1.0pt]{1-10}
\end{tabular}
\caption{Word similarity evaluation on Multi-SimLex (Spearman's $\rho$).}
\label{tab:ws_full}
\end{table*}

\begin{table*}[!t]
\centering
\small
\begin{tabular}{lccccccccccc}
\cmidrule[1.0pt]{1-10}
model$\downarrow$, lang.$\rightarrow$  & \es & \ar & \ru & avg. \\
\cmidrule[1.0pt]{1-10}
BERT-CLS & .526 & .308 & .470 & .435 \\
BERT-mp & .599 & .455 & .552 & .535 \\
\rowcolor{blue!10}
+ Mirror & \underline{.709} & \textbf{.669} & \underline{.673} & \textbf{.684} \\
\cmidrule[1.0pt]{1-10}
mBERT-CLS & .421 & .326 & .430 & .392 \\
mBERT-mp & .610 & .447 & .616 & .558 \\
\rowcolor{blue!10}
 + Mirror & \textbf{.755} & \underline{.594} & \textbf{.692} & \underline{.680}\\
\cmidrule[1.0pt]{1-10}
\end{tabular}
\caption{Full Spanish, Arabic and Russian STS evaluation. Spearman's $\rho$ correlation reported.}
\label{tab:sts_multillingual_full}
\end{table*}

\begin{table*}[] 
\centering
\small
\begin{tabular}{lccccccccccc}
\cmidrule[1.0pt]{1-10}
lang.$\rightarrow$ & \en-\fr & \en-\zh & \en-\he & \fr-\zh & \fr-\he & \zh-\he & avg. \\
\cmidrule[1.0pt]{1-10}
mBERT-CLS & .059 & .053 & .032 & .042 & .024 & .050 & .043 \\
mBERT-mp & .163 & .118 & .071 & .142 & .104 & .010 & .101 \\
\rowcolor{blue!10}
+ Mirror  & \textbf{.454} & \textbf{.385} & \textbf{.133} & \textbf{.465} & \textbf{.163} & \textbf{.179} & \textbf{.297} \\
\cmidrule[1.0pt]{1-10}
\end{tabular}
\caption{Full cross-lingual word similarity evaluation on Multi-SimLex (Spearman's $\rho$). }
\label{tab:ws_xling_full}
\end{table*}

\begin{table*}[] 
\centering
\small
\begin{tabular}{lcccccccccccc}
\cmidrule[1.0pt]{1-10}
lang.$\rightarrow$  & \en-\fr & \en-\ita & \en-\ru & \en-\tr & \ita-\fr & \ru-\fr & avg. \\
\cmidrule[1.0pt]{1-10}
BERT-CLS & .045 & .049 & .108 & .109 & .046 & .068 & .071 \\
BERT-mp & .014 & .112 & .154 & .150 & .025 & .018 & .079 \\
\rowcolor{blue!10}
+ Mirror  & \textbf{.458} & \textbf{.378} & \textbf{.336} & \textbf{.289} & \textbf{.417} & \textbf{.345} & \textbf{.371} \\
\cmidrule[1.0pt]{1-10}
\end{tabular}
\caption{Full Bilingual Lexicon Induction results (accuracy reported). ``\en-\fr'' means en mapped to \fr.}
\label{tab:bli_full}
\end{table*}

\begin{table*}[!htbp]
\centering
\small
\begin{tabular}{lccccccclccc}
\cmidrule[1.0pt]{1-10}
model configuration$\downarrow$, dataset$\rightarrow$ & STS12 & STS13 & STS14 & STS15 & STS16 & STS-b & SICK-R & avg.\\
\cmidrule[1.0pt]{1-10}
BERT + Mirror & .674 & .796 & .713 & .814 & .743 & .764 & .703 & .744 \\
\hdashline
- dropout & .646 & .770 & .691 & .800 & .726 & .745 & .701 & .726$_{\downarrow .018}$  \\
- random span masking & .641 & .775 & .684 & .777 & .737 & .749 & .658 & .717$_{\downarrow .027}$ \\
- dropout \& random span masking & .587 & .695 & .617 & .688 & .683 & .674 & .614 & .651$_{\downarrow .093}$ \\
\cmidrule[1.0pt]{1-10}
RoBERTa + Mirror & .648 & .819 & .732 & .798 & .780 & .787 & \underline{.706} & .753 \\
\hdashline
 - dropout & .619  & .795 & .706 & .802 & .777 & .727 & .698 & .732$_{\downarrow .021}$ \\
 - random span masking & .616 & .786 & .689 & .766 & .743 & .756 & .663 & .717$_{\downarrow .036}$ \\
 - dropout \& random span masking & .562 & .730 & .643 & .744 & .752 & .708 & .638 & .682$_{\downarrow .071}$ \\
\cmidrule[1.0pt]{1-10}
\end{tabular}
\caption{Full table for the synergistic effect of dropout and random span masking in sentence similarity tasks.}
\label{tab:synersitic_full}
\end{table*}

\section{Number of Model Parameters}\label{sec:appendix_number_param}
All BERT/RoBERTa models in this paper have $\approx$110M parameters.

\section{Hyperparameter Optimisation}\label{sec:hyperparameters}
\Cref{Table:search_space} lists the hyperparameter search space. Note that the chosen hyperparameters yield the overall best performance, but might be suboptimal on any single setting (e.g. different base model). 

\begin{table*}[!ht] 
\small
\centering
\begin{tabular}{lr}
\toprule
hyperparameters & search space \\
\midrule
learning rate & \{\texttt{5e-5}, \texttt{2e-5}$^\ast$, \texttt{1e-5}\} \\
batch size & \{100, 200$^\ast$, 300\} \\
training epochs & \{1$^\ast$, 2$^\ast$, 3, 5\}\\
$\tau$ in \Cref{eq:infonce} & \{0.03, 0.04$^\ast$, 0.05, 0.07, 0.1, 0.2$^\ast$, 0.3\}\\
\bottomrule
\end{tabular}
\caption{Hyperparameters along with their search grid. $^\ast$ marks the values used to obtain the reported results. The hparams are not always optimal in every setting but generally performs (close to) the best.}
\label{Table:search_space}
\end{table*}

\section{Software and Hardware Dependencies}
All our experiments are implemented using PyTorch 1.7.0 and \url{huggingface.co} transformers 4.4.2, with Automatic Mixed Precision (AMP)\footnote{\url{https://pytorch.org/docs/stable/amp.html}} turned on during training. Please refer to the GitHub repo for details. The hardware we use is listed in \Cref{Table:hardware}. 

\begin{table*}[h] 
\small
\centering
\begin{tabular}{lr}
\toprule
hardware & specification \\
\midrule
RAM & 128 GB \\
CPU & AMD Ryzen 9 3900x 12-core processor × 24  \\
 GPU & NVIDIA GeForce RTX 2080 Ti (11 GB) $\times$ 2\\
\bottomrule
\end{tabular}
\caption{Hardware specifications of the used machine. When encountering out-of-memoery error, we also used a second server with two NVIDIA GeForce RTX 3090 (24 GB).}
\label{Table:hardware}
\end{table*}

\end{document}